\documentclass[sigconf]{acmart}
\usepackage{algorithmic}
\usepackage{algorithm, multirow}
\AtBeginDocument{%
  }

\copyrightyear{2025}
\acmYear{2025}
\setcopyright{acmlicensed}\acmConference[MMAsia '25]{ACM Multimedia Asia}{December 9--12, 2025}{Kuala Lumpur, Malaysia}
\acmBooktitle{ACM Multimedia Asia (MMAsia '25), December 9--12, 2025, Kuala Lumpur, Malaysia}
\acmDOI{10.1145/3743093.3771008}
\acmISBN{979-8-4007-2005-5/2025/12}

\begin{document}

\title{MCIHN: A Hybrid Network Model Based on Multi-path Cross-modal Interaction for Multimodal Emotion Recognition}

\author{Haoyang Zhang}

\orcid{0009-0009-1581-3147}
\affiliation{%
	\institution{Chongqing University of Posts and Telecommunications}
	\city{Chongqing}
	\country{China}}
\email{2021214892@stu.cqupt.edu.cn}

\author{Zhou Yang}
\authornote{Corresponding author.}
\affiliation{%
	\institution{Xi’an Jiaotong University}
	\city{Xi’an}
	\country{China}}
\email{yzhoul392@gmail.com}

\author{Ke Sun}
\affiliation{%
	\institution{University of New South Wales}
	\city{Sydney}
	\country{Australia}}
	\email{sk3270148848@163.com}

\author{Yucai Pang}
\affiliation{%
	\institution{Chongqing University of Posts and Telecommunications}
	\city{Chongqing}
	\country{China}}
	\email{pangyc@cqupt.edu.cn}

\author{Guoliang Xu}
\affiliation{%
	\institution{Chongqing University of Posts and Telecommunications}
	\city{Chongqing}
	\country{China}}
	\email{xugl@cqupt.edu.cn}

\renewcommand{\shortauthors}{Haoyang Zhang et al.}


\begin{abstract}
  Multimodal emotion recognition is crucial for future human-computer interaction. However, accurate emotion recognition still faces significant challenges due to differences between different modalities and the difficulty of characterizing unimodal emotional information. To solve these problems, a hybrid network model based on multipath cross-modal interaction (MCIHN) is proposed. First, adversarial autoencoders (AAE) are constructed separately for each modality. The AAE learns discriminative emotion features and reconstructs the features through a decoder to obtain more discriminative information about the emotion classes. Then, the latent codes from the AAE of different modalities are fed into a predefined Cross-modal Gate Mechanism model (CGMM) to reduce the discrepancy between modalities, establish the emotional relationship between interacting modalities, and generate the interaction features between different modalities. Multimodal fusion using the Feature Fusion module (FFM) for better emotion recognition. Experiments were conducted on publicly available SIMS and MOSI datasets, demonstrating that MCIHN achieves superior performance.
\end{abstract}



\maketitle

\section{Introduction}

With the booming development of information technology, affective computing \cite{tao2009affective} has become an important research task in the field of computer science. In recent years, with the rapid development of deep learning, people try to make machines have the ability to analyze human emotions through machine learning or deep learning methods. At the same time, accurate recognition of human emotions is urgent because of the need for emotion recognition in human-computer interaction, emotion-aware recommendation, and other areas.

Emotion recognition is a complex task involving spatial fusion modeling of multi-dimensional features \cite{peng2023fine,wang2023multimodal,wang2023cross}, because human emotion expression is highly complex. Typically, people express emotions through voice, facial expressions, body gestures, etc \cite{fu2024hybrid,sun2023using,yang2023behavioral}. Human emotional expressions are co-constructed by different modalities and different modalities are highly correlated with the same emotional expression \cite{zhang2023deep}. In essence, emotion recognition is more appropriately viewed as a multimodal emotion recognition problem. Considering that the emotional characteristics and expressive strengths of different modal information are not consistent \cite{sun2024fine}, this makes the multimodal emotion recognition task highly complex.

Autoencoder (AE) is an unsupervised learning model that has been successfully applied to emotion recognition. AE is very powerful in learning salient representations and can lead to a significant improvement in emotion recognition performance \cite{zheng2023multi}. Adversarial Autoencoder (AAE) is a probabilistic model that improves on AE as a generative model. This makes AAE more powerful in learning more descriptive features than traditional AE or even variational autoencoders (VAE) \cite{makhzani2018unsupervised}.

The previous research has focused on cross-modal feature interaction and multimodal feature fusion. In terms of cross-modal feature interaction, Rahman et al \cite{rahman2020integrating} proposed to improve the fine-tuning process of the model using multimodal fitness gates to enhance modal interaction. Hazarika et al \cite{hazarika2020misa} proposed combining different losses to capture cross-modal commonalities to reduce modal gaps. Combining the previous work, we propose cross-modal gate mechanism in this paper to enable cross-modal information interaction and capture modal commonality information. In terms of multimodal feature fusion, Mirsamadi et al \cite{mirsamadi2017automatic} proposed feature pooling strategies with localized attention that can focus attention on more emotionally salient regions. Li et al \cite{li2019towards} use a multi-head self-attention mechanism to model the relative dependencies between elements. Given that previous work has experimentally demonstrated the effectiveness of multi-head attention mechanisms in feature selection of key emotional information and in learning long-term dependencies of sequences. We use  several multi-head attention mechanisms interacting for fusion. To solve the problems of ignoring the influence of affective interactions between different modal data and the limited effect of in-depth processing of modal feature information in a way that characterizes the modal features.  this paper proposes the MCIHN model. The overall structure of the model MCIHN is shown in Fig.1(a). The contributions of this paper are as follows:

\begin{itemize}
	\setlength{\itemsep}{-1pt}
	\setlength{\parsep}{0pt}
	\setlength{\parskip}{0pt}
	\vspace{-2mm}
	\item[1.]
	We construct AAE models to process different modal features separately. The features are reconstructed and fed back by the decoder so that the latent codes contain more discriminatory emotional information, with both temporal and spatial information to increase the expressiveness of the latent codes.
	
	\item[2.]	
	Propose a predefined Cross-modal Gate Mechanism model to reduce the discrepancy between the modalities, establish affective relationships between the interactive modalities, and generate interaction features, thereby ensuring the effective processing of complex input combinations.
	
	\item[3.] Experiments were conducted on the Chinese dataset SIMS and the English dataset MOSI. The MCIHN model was evaluated on a multi-sentiment classification task, and the results demonstrate that the model performs better on these two datasets.
\end{itemize} 

\begin{figure*}[htbp]
	\includegraphics[width=1.0\linewidth] {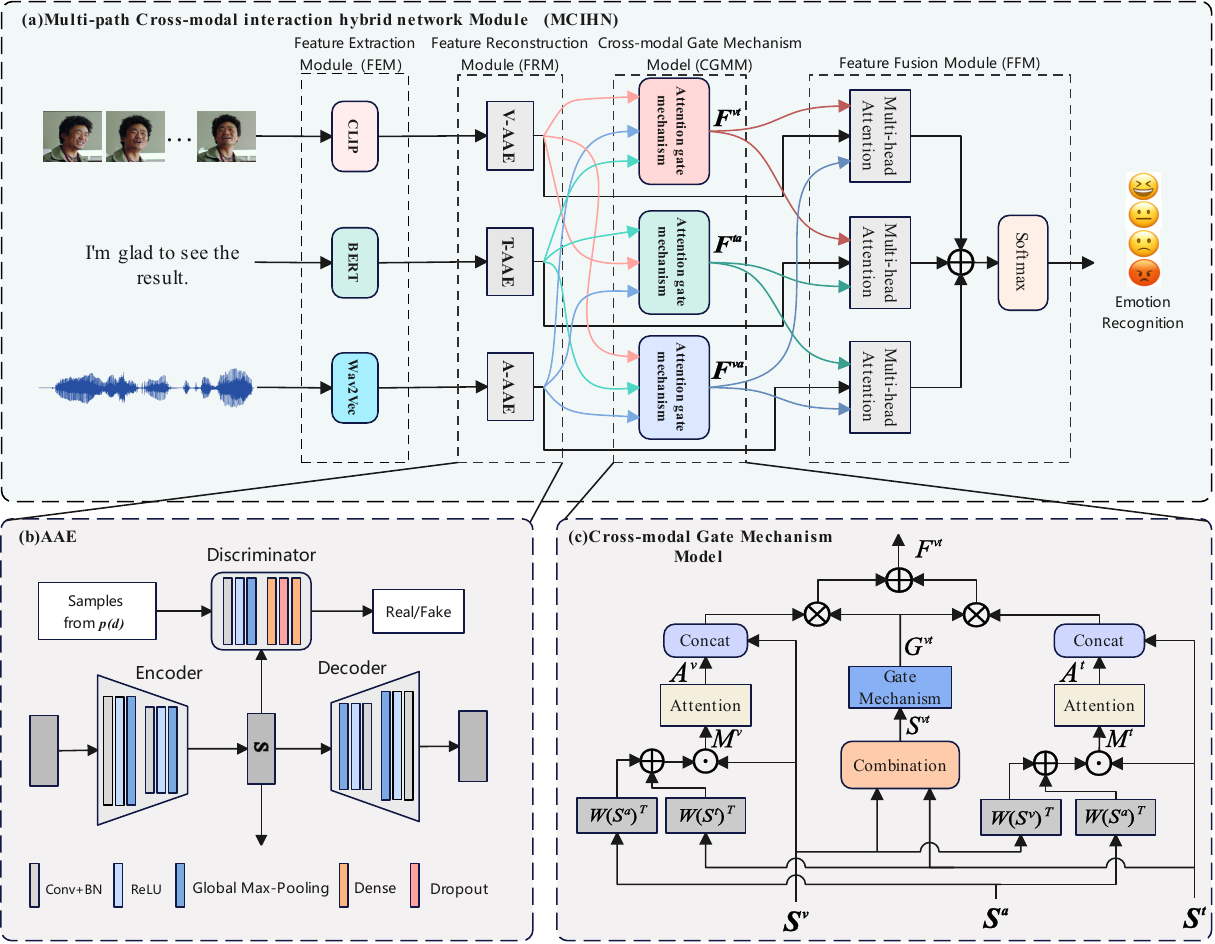}
	\caption{Framework of the MCIHN model.}
	\label{fig-model}
\end{figure*}

\section{Related Work}
With the widespread emergence of multimodal task scenarios, multimodal emotion recognition tasks have become an important research topic for integrating visual, text, audio and even brainwave data. In addition to feature representation of unimodal data, feature fusion methods and emotion recognition methods for multimodal data are also hot topics in current research. The following section describes the related research in detail.

A large amount of work in recent years has shown that traditional multimodal information fusion methods include early fusion, late fusion, and model-level fusion \cite{zhang2023deep}. Early fusion, also known as feature-level fusion, is an approach that concatenates features extracted from a single modality into a complete feature vector, which is then passed through a classifier for sentiment classification. However, this early fusion approach does not take into account the time scale and cannot capture the association between different modalities. Later fusion, also known as decision-level fusion, follows certain mathematical rules, such as “sum”, “product”, “mean”, “maxima ”, “minimum value”, etc., so that the results obtained from various modalities are combined into a final result. The advantage of late fusion is that each modality can be individually selected with an appropriate sentiment classifier. However, the late fusion method considers that the different modalities are still independent from each other, so it cannot reveal the relationship between different modalities. Model-level fusion is widely used in emotion tasks, which can consider the interrelationships among different modalities and reduce the need for temporal synchronization of these different modalities. 

Recently, a large number of multimodal modeling algorithms have been proposed, e.g., MESM \cite{dai2021multimodal}, MMAN \cite{pan2020multi}, and so on.Multimodal End-to-End Sparse Model (MESM) for sentiment classification. For the pre-trained VGG16 model to complete the extraction of raw audio and image features, then the temporal information is modeled using transformer, and the textual modality is encoded directly into the word sequence using transformer. However, after modeling the temporal information the representation of the spatio-temporal dimension of the features needs to be better extracted, and the unimodal features need to be better processed in terms of temporal order and feature points. Pan et al \cite{pan2020multi} designed a hybrid fusion strategy for multimodal emotion recognition called Multimodal Attention Network (MMAN), where MMAN consists of an early fused multiattention cLSTM and three independent unimodal modules, which are then used in combination with an FC layer and a softmax layer for later fusion.In addition, Mai et al \cite{mai2020multi} proposed a hybrid contrast learning approach for multimodal sentiment analysis with a three-modal feature representation, where COVAREP \cite{degottex2014covarep} is taken for audio to extract typical LLDs, a set of hand-extracted features is used for vision, and pre-trained Glove is used for text to achieve advanced textual features. The inter-sample and inter-class correlations are then captured by intra- and inter-modal contrast learning and semi-contrast learning. None of these approaches are good enough for modal fusion, and the use of different types of attentional mechanisms has been proposed for models with more complex combinations of inputs. However, issues such as the lack of cross-modal affective influence in affective interactions between multiple modalities remain to be addressed.

Based on these issues, this paper proposes a novel feature fusion model, which focuses on intra-modal inter-modal and interactive inter-modal affective differences as well as affective relationships during interaction in the fusion process.

\section{Methodology}
\subsection{Problem Formalization}
As shown in Fig.1(a), CLIP, BERT, and Wav2vec2 are selected as feature extractors for visual, text, and audio modalities, respectively. The sequences extracted by the feature extractors for each modality are denoted as $X_v\in R^{T_v\times D_v }$,$ X_t\in R^{T_t\times D_t }$, $X_a\in R^{T_a\times D_a }$, where $T_v$, $T_t$, $T_a$ denote the lengths of the visual, text, and audio sequences, respectively, and $D_v$, $D_t$, $D_a$ denote their dimensions. The output of the model is denoted by $y^p$, and the real emotion label is denoted by $y^g$.

\subsection{Feature Extraction Module}
CLIP, BERT, Wav2Vec2 are chosen as the visual, text, speech feature extractors and the features can be represented as:
\begin{equation}
	X_v=CLIP(Video)
\end{equation}
\begin{equation}
	X_t=BERT(Text)
\end{equation}
\begin{equation}
	X_a=Wav2Vec(Audio)
\end{equation}
Video, Text and Audio are unprocessed visual, text and voice inputs respectively.

\subsection{Feature Reconstruction Module}
As shown in Fig. 1(b), the discriminator forces the autoencoder to generate the latent representation $S$ and be more regular in distribution by the statistical properties of the given prior distribution $p\left(d\right)$. The adversarial part is attached to the hidden code S. Here, the encoder is also the generator of the adversarial network. The decoder part updates S by reconstructing the  representations so that S contains a greater amount of affective representational detail capable of characterizing the original input. For the input data $X$, the overall model is divided into two phases: 

(1) Reconstruction phase : autoencoder updates the encoder $E_\varepsilon$ and the decoder $D_\delta$ and minimizes the reconstruction error by encoding $X$ as a latent representation $S$. The objective function of the autoencoder is defined as:
\begin{equation}
	\mathcal{L}_{AE}\left( {X,D_{\delta}\left( {E_{\varepsilon}(X)} \right)} \right) = \left\| {X - \hat{X}} \right\|_{2}^{2}
\end{equation}

(2) Regularisation phase: the adversarial network first updates the discriminator $D_\omega$ to distinguish between samples from the prior distribution $p\left(d\right)=\mathcal{N}(d;0,I)$ and samples generated by computing the latent code S using the encoder, and then updates the encoder. The update is achieved by keeping the weights and biases of the discriminator unchanged and back-propagating the biases to $E_\varepsilon$ and then updating its weights and bias values. $P_X$ is the data distribution. We define the updated encoder as $E_\alpha^\ast$. The objective function of the discriminator $D_\omega$ is defined as:
\begin{equation}
	\begin{aligned}
		\mathcal{L}_{disc} = \underset{\omega}{max}{({E_{d\sim P_{d}}\left\lbrack {\log\left( D_{\omega}(d) \right)} \right\rbrack})} \\
		+ E_{X\sim P_{X}}\left\lbrack log\left( 1 - D_{\omega}\left( E_{\varepsilon}\left( \left. X \right) \right) \right) \right\rbrack \right. \\
	\end{aligned}
\end{equation}

This stage is performed sequentially, starting with the reconstruction stage update, followed by regularisation. The final latent code is represented as $S=E_\alpha^\ast(X)$. The latent codes obtained in the three different modal paths of vision, text, and audio are $S^v$, $S^t$, and $S^a$ respectively.

\subsection{Cross-modal Gate Mechanism Model}
As shown in Fig.1(c), taking the modal correlation features of visual and text as an example, the gate mechanism first generates the correlation features between the two modalities. It then reduces the affective difference between the associated features and the core features by minimizing the Maximum Mean Deviation (MMD) between them, thus establishing the affective correlation between the different modalities. The mechanism of Attention affective transmission, dominated by the visual path, can be formulated as follows:
\begin{equation}		
	M^{v} = S^{v}\left\lbrack W\left( S^{a} \right)^{T} + {W\left( S^{t} \right)}^{T} \right\rbrack
\end{equation}
\begin{equation}
	\alpha_{1}\left( {i,} \right) = Softmax\left( M^{v}(i,) \right)
\end{equation}
\begin{equation}
	A^{v} = \alpha_{1} * M^{v}
\end{equation}
where $W\in R^{d\times d}$ is the trainable matrix; $M^v$ is the results of the interaction matching matrix computation; $\alpha\left( {i,} \right)$ is the attention score.

The visual and textual modal information is combined, and then the gate mechanism is used to filter the information containing these two modalities. The expression is as follows:
\begin{equation}	
	S^{vt} = \left\lbrack {S^{v},S^{t},S^{v} - S^{t},S^{v} \cdot S^{t}} \right\rbrack\left( W_{vt} \right) + \left( b^{vt} \right)^{T}  		
\end{equation}
\begin{equation}	
	G^{vt} = ReLU\left( S^{vt} \right) 	
\end{equation}
where $W_{vt}\in R^{d\times d}$ and $b^{vt}\in R^{d\times1}$ are trainable parameters and $G^{vt}$ denotes the output of the gating mechanism ReLU.

The combination of the Attention affective transfer mechanism and the cross-modal gating mechanism is defined as a joint representation of visual and textual modalities, expressed as:
\begin{equation}
	F^{vt} = G^{vt}\left\lbrack {A^{v},S^{v}} \right\rbrack + G^{vt}\left\lbrack A^{t},S^{t} \right\rbrack
\end{equation}

A feature adaptive approach \cite{huang2017cross} is used to reduce the Maximum Mean Discrepancy (MMD) between the associated features of a modality and its core features. Use $h_k^2(S,F)$ to denote the MMD between the core molar distribution ${I^{core}}$ of the core modality $S^{core}$ and the molar correlation distribution ${I^{other}}$ of the other two modalities $S^{other}$. In this way, the MMD between the visual modality as the core modality, for example, and the other modalities can be expressed as the square formula in the reproducing kernel Hilbert space $x_k$:
\begin{equation}
	h_{k}^{2}\left( {S^{v-core},S^{other}} \right) \triangleq {\parallel E_{H}\left\lbrack {\xi\left( I_{v} \right)} \right\rbrack - E_{H}\left\lbrack {\xi\left( I_{ta} \right)} \right\rbrack \parallel}_{x_{k}}^{2}
\end{equation}

Here $\xi$ is a fully connected layer. $\Gamma_{k}\left( S^{core} \right)$ is the average representation of the distribution $S^{core}$ in $x_{k}$ ,for all f, needs to satisfy the following conditions:
\begin{equation}
	E_{X \sim H}f(X) = \left\lbrack F(X),\Gamma_{k}\left( S^{core} \right) \right\rbrack_{x_{k}}^{2}
\end{equation}
The cross-modal correlation feature adaptive loss function can therefore be expressed as:
\begin{equation}
	\mathcal{L}\left( {Adp} \right) = {\sum\limits_{v}{\sum\limits_{i = 1}^{K}{h_{k}^{2}\left( S_{i}^{v - core},S_{i}^{ta} \right)}}}
\end{equation}
By minimizing $\mathcal{L}\left( {Adp} \right)$ can effectively reduce the difference in emotion between the core modality and the other modalities, thus enabling cross-modal emotion transfer.

\subsection{Feature Fusion Module}
\par As shown in Fig. 1(a), we design the multi-head attention mechanism based on different associated features and generate new joint representations to obtain $F^{v-ta}$, $F^{t-va}$, and $F^{a-vt}$ by using the cascading features of other auxiliary modalities and the main modality as weights. Taking $F^{v-ta}$ as an example it can be specifically formulated as:

\begin{equation}	
	\begin{split}
		{head}_{i} &= Attention\left( {Q\left( F^{vt} \right),K\left( F^{va} \right),V\left( F^{v} \right)} \right)    \\
		&= softmax\left( \frac{F^{vt}W_{i}^{Q}\left( F^{va}W_{i}^{K} \right)}{\sqrt{d_{k}/h}} \right)F^{v}W_{i}^{V}     \\
	\end{split}
\end{equation}

\begin{equation}	
	F^{v - ta} = Concat\left( {head}_{1},\ldots,{head}_{h} \right)W^{O}
\end{equation}

Finally, the output of the model can be expressed as:
\begin{equation}
	y^{p} = Softmax\left( Linear\left( Add\left( F^{v - ta},F^{t - va},F^{a - vt} \right) \right) \right)
\end{equation}

The mean absolute error is used to calculate the classification loss and the number of training samples is $N$. The loss function for the predicted output and the true sentiment output is expressed as:
\begin{equation}
	\mathcal{L}\left( {Mul} \right) = \frac{1}{N}{\sum\limits_{i = 1}^{N}\left| {y_{i}^{p} - y_{i}^{g}} \right|}
\end{equation}

We jointly train the cross-modal association feature adaptive loss and the classification loss. The joint training loss can be expressed as :
\begin{equation}
	\mathcal{L}\left( {Combined} \right) = \mathcal{L}\left( {Adp} \right) + \mathcal{L}\left( {Mul} \right)
\end{equation}

\subsection{Learning Algorithms and Time Complexity}
\begin{algorithm}[t]
	\caption{Training Algorithm for MCIHN Model}
	\label{alg:MCIHN}
	\textbf{Input:} Unimodal inputs: Visual $X_v$, Text $X_t$, Audio $X_a$; Ground-truth labels $\mathbf{y}^g$. \\
	\textbf{Output:} Predicted labels $\mathbf{y}^p$; Optimal parameter set $\theta_{\text{opt}}$.
	\begin{algorithmic}[1]
		\STATE Initialize parameters for AAE encoders $E_\epsilon$, decoders $D_\delta$, discriminator $D_\omega$, CGMM, and FFM
		\STATE Set max epochs $L$, batch size $K$
		\FOR{epoch $= 1 : L$}
		\FOR{batch $i = 1 : K$}
		\STATE \textbf{Feature Extraction:}
		\STATE Extract features via (1)-(3): 
		\STATE $X_v \gets \text{CLIP}(Video),\ X_t \gets \text{BERT}(Text),\ X_a \gets \text{Wav2Vec}(Audio)$
		
		\STATE \textbf{Feature Reconstruction via AAE:}
		\STATE Encode latent codes: $S^v \gets E_\epsilon^v(X_v),\ S^t \gets E_\epsilon^t(X_t),\ S^a \gets E_\epsilon^a(X_a)$
		\STATE Reconstruct outputs: $\hat{X}_v \gets D_\delta^v(S^v),\ \hat{X}_t \gets D_\delta^t(S^t),\ \hat{X}_a \gets D_\delta^a(S^a)$
		\STATE Compute $\mathcal{L}_{AE}$ for each modality using (4)
		\STATE Update $E_\epsilon$ and $D_\delta$ to minimize $\mathcal{L}_{AE}$
		\STATE Compute $\mathcal{L}_{disc}$ via (5); Update $D_\omega$ and $E_\epsilon$ adversarially
		
		\STATE \textbf{Cross-modal Gate Mechanism:}
		\STATE Compute interaction features $F^{vt}, F^{va}, F^{ta}$ via (6)-(11)
		\STATE Calculate MMD loss $\mathcal{L}(Adp)$ using (12)-(14)
		
		\STATE \textbf{Feature Fusion Module:}
		\STATE Generate fused features $F^{v-ta}, F^{t-va}, F^{a-vt}$ via (15)-(16)
		\STATE Compute final prediction $\mathbf{y}^p$ using (17)
		\STATE Calculate classification loss $\mathcal{L}(Mul)$ via (18)
		
		\STATE \textbf{Joint Optimization:}
		\STATE Compute total loss $\mathcal{L}(Combined) \gets \mathcal{L}(Adp) + \mathcal{L}(Mul)$ via (19)
		\STATE Backpropagate and update all parameters to minimize $\mathcal{L}(Combined)$
		\ENDFOR
		\ENDFOR
	\end{algorithmic}
\end{algorithm}

The algorithmic time complexity of the MCIHN model proposed in this paper is mainly composed of three core modules: adversarial autoencoder (AAE), cross-modal gate mechanism (CGMM), and multi-head attentional feature fusion module (FFM).The AAE module needs to independently process the features of the three modalities of vision, text, and speech, and assuming that the time complexity of each AAE is \(O(N \cdot d^2)\) (\(N\) is the number of samples, \(d\) is the feature dimension), then the total complexity of the three modalities is \(3 \cdot O(N \cdot d^2)\); the CGMM module calculates the correlation matrix and applies the gating mechanism through the cross-modal interaction, and the time complexity of its two-by-two modal interaction is \(O(d^2)\), and the total complexity under the three-modal combination is \(3 \cdot O(d^ 2)\); the FFM module uses a multi-head attention mechanism with a time complexity of \(O(n^2 \cdot d)\) (\(n\) is the sequence length). The overall time complexity can be summarized as \(O(N \cdot d^2 + k \cdot d^2 + n^2 \cdot d)\) (\(k\) is the number of modal combinations).

\section{Experiments and Analysis}

\subsection{Datasets and Settings}

\begin{table}[!h]
	\centering
	
	\scalebox{1}{	
		\begin{tabular}{c|c c c c}\toprule[1.7pt]
			
			Dataset & Train & Test & Valid & All  \\ \hline \hline
			SIMS & $742/207/419$  & $248/69/140$ & $248/69/139$  & $2281$ \\ \hline
			CMU-MOSI & $552/53/679$ & $379/30/277$ & $92/13/124$  & $2199$ \\ \hline \hline
			\toprule[1.3pt]
		\end{tabular}
	}
	\caption{The three numbers for each split denote the number of the samples with negative ($< 0$), neutral ($= 0$), and positive ($> 0$) sentiment, respectively.}
	\label{table_dataset}
\end{table}

\textbf{CH-SMIS:} This dataset is a Chinese benchmark dataset for multimodal emotion recognition tasks \cite{yu2020ch}. It consists of 2281 refined video segments with multimodal and individual modal labels. Sentiment scores are marked as positive, weakly positive, neutral, negative, and weakly negative. In this paper, positive and weakly positive were marked as 1, negative and weakly negative as -1, and neutral as 0.

\textbf{MOSI:} This dataset is currently a widely used benchmark dataset for MSA tasks \cite{zadeh2016multimodal}. It consists of 2199 short monologue video clips separated from 93 YouTube movie review videos. Sentiment scores range from -1 to +1 and are labeled as strongly negative, negative, weakly negative, neutral, weakly positive, positive, and strongly positive.

\begin{table*}[!h]
	\centering
	\resizebox{\linewidth}{!}{
		\scalebox{0.7}{	
			\begin{tabular}{c|c c c c c c|c c c c c}\toprule[0.8pt]
				&&&{CH-SIMS}&&&     &  &  &CMU-MOSI  &  &        \\ \hline \hline
				Methods   & Acc-2 & F1 & Acc-3 & Acc-5 & MAE & Corr                               & Acc-2 & F1 & Acc-7 & MAE & Corr                              \\ \hline \hline
				TFN       & $78.3$  & $78.6$ & $65.1$ & $39.3$ & $0.432$ & $0.591$                & $80.8$  & $80.7$ & $34.9$ & $0.901$ & $0.698$           \\ \hline
				LMF       & $77.7$  & $77.8$ & $64.6$ & $40.5$ & $0.441$ & $0.576$                & $82.5$  & $82.4$ & $33.2$ & $0.917$ & $0.695$                   \\ \hline
				MulT      & $78.2$  & $78.5$ & $65.7$ & $40.0$ & $0.442$ & $0.581$                & $83.0$  & $82.8$ & $40.0$ & $0.871$ & $0.698$                    \\ \hline
				Self-MM   & $80.0$  & $80.4$ & $65.4$ & $41.5$ & $0.425$ & $0.595$                & $83.4$  & $83.3$ & $46.7$ & $0.708$ & $0.796$         \\ \hline
				EMT       & $80.1$  & $80.1$ & $67.4$ & $43.5$ & $0.396$ & $0.623$                & $83.3$  & $83.2$ & $47.4$ & $0.705$ & $\mathbf{0.798}$     \\ \hline  
				CMHFM     & $82.2$  & $82.3$ & $66.8$ & $41.8$ & $0.408$ & $0.649$       & $81.7$  & $81.5$ & $37.5$ & $0.822$ & $0.723$    
				
				\\ \hline

				MCIHN   & $\mathbf{82.8}$  & $\mathbf{82.6}$ & $\mathbf{68.8}$ & $\mathbf{45.2}$ & $\mathbf{0.391}$ & $\mathbf{0.653}$             & $\mathbf{84.4}$  & $\mathbf{84.1}$ & $\mathbf{48.2}$ & $\mathbf{0.701}$ & $0.796$  \\ \hline
				\toprule[0.8pt]
			\end{tabular}
		}
	}
	\caption{Baseline Algorithms Results on CH-SIMS and CMU-MOSI}
	\label{table_dataset1}
\end{table*}

\begin{table*}[!h]
	\centering
	\resizebox{\linewidth}{!}{
		\scalebox{0.7}{	
			\begin{tabular}{c|c c c c c c|c c c c c}\toprule[0.8pt]
				&&&{CH-SIMS}&&&     &  &  &CMU-MOSI  &  &        \\ \hline \hline
				Methods   & Acc-2 & F1 & Acc-3 & Acc-5 & MAE & Corr                                       & Acc-2 & F1 & Acc-7 & MAE  & Corr                                          \\ \hline \hline
				
				-VT  & $78.7$  & $78.5$  & $65.4$  & $41.5$  & $0.428$  & $0.596$                    & $80.9$ & $80.6$ & $35.7$ & $0.734$ & $0.732$ \\ \hline
				-VA  & $77.3$  & $77.2$  & $63.7$  & $40.7$  & $0.449$  & $0.586$                    & $79.9$ & $79.9$ & $33.9$ & $0.798$ & $0.729$ \\ \hline
				-TA  & $77.6$  & $77.6$  & $64.1$  & $40.1$  & $0.442$  & $0.581$                    & $80.1$ & $80.0$ & $33.4$ & $0.757$ & $0.739$ \\ \hline
				MCIHN-1   & $78.8$  & $78.8$  & $63.9$  & $40.9$  & $0.433$  & $0.586$                    & $80.3$ & $79.6$ & $33.6$ & $0.781$ & $0.732$  \\ \hline
				MCIHN-2   & $77.8$  & $75.6$  & $63.2$  & $39.8$  & $0.446$  & $0.574$                    & $79.6$ & $78.9$ & $32.7$ & $0.774$ & $0.715$  \\ \hline		
				MCIHN   & $\mathbf{82.8}$  & $\mathbf{82.6}$ & $\mathbf{68.8}$ & $\mathbf{45.2}$ & $\mathbf{0.391}$ & $\mathbf{0.653}$             & $\mathbf{84.4}$  & $\mathbf{84.1}$ & $\mathbf{48.2}$ & $\mathbf{0.701}$ & $\mathbf{0.796}$  \\ \hline
				\toprule[0.8pt]
			\end{tabular}
		}
	}
	\caption{Ablation Experiment Results on CH-SIMS and CMU-MOSI}
	\label{table_dataset2}
\end{table*}

\subsection{Evaluation Metrics}
Aligning with existing literature \cite{sun2023efficient}, since multimodal emotion recognition is a regression task, we use Mean Absolute Error (MAE) and Pearson correlation coefficient (Corr) for evaluation. Researchers also convert the continuous score into different discrete categories and report classification accuracy. Following previous work, we record binary accuracy (Acc-2), seven-class accuracy (Acc-7), and F1-score on CMU-MOSI. On CH-SIMS, we report binary accuracy (Acc-2), three-class accuracy (Acc-3), five-class accuracy (Acc-5), and F1-score. For all metrics but MAE, higher values indicate better performance.

\subsection{Experimental Setup}
In the experiments, CLIP, BERT, and Wav2Vec are used as visual, textual, and speech feature extractors.

\textbf{For Vision Modality.} These two datasets are processed using the CLIP-based ViT-B-32 model. This pre-trained model uses the Transformer architecture, which is capable of handling large-scale image features and generating high-quality features.

\textbf{For Text Modality.} Transformer-based pre-trained language models have achieved state-of-the-art performances on a wide range of tasks in natural language processing. In agreement with recent works \cite{sun2023efficient}, we employ the pre-trained BERT model from the open-source Transformers library to encode raw text. Specifically, we use bert-base-uncased model for CMU-MOSI and bert-base-chinese model for CH-SIMS.

\textbf{For Audio Modality.} The Wav2vec2 model has achieved state-of-the-art performance in a variety of speech tasks. We use the pre-trained Wav2vec2 model from the open source Transformers library to encode speech. Specifically, we use the wav2vec2-large-xlsr-53-english model in CMU-MOSI, and the wav2vec2-large-xlsr-53-english-zh-cn-gpt Chinese model in CH-SIMS.

The pre-processed visual, text, and speech data sizes are 512, 768, and 512, respectively, and the sequence lengths are set to 10, 36, and 128, respectively. The output features of the same uniform size are then set to 256 and the sequence length to 32, respectively, by different parameter settings in the AAE module, and the sizes are kept unchanged in the multimodal interaction fusion module. Throughout the experiments, we select Adam as the optimizer to train the model and use the early stopping technique to determine the number of iterations. For BERT, we set the learning rate value as 2e-5, and for other modules, the learning rate is set as 1e-3 to determine the optimal value through experiments. The number of batch size is set as 32 and the dropout rate is 0.5. The hyperparameters in the model are tuned within the range [0.001,0.01,0.1,0.3,0.5,0.7,0.9,0.99] to obtain the optimal values.

\subsection{Baseline Algorithms}
We compared the MCIHN model with the general baseline algorithm and the latest algorithms.

\textbf{TFN.} Tensor Fusion Network (TFN) \cite{zadeh2017tensor} introduces a three-fold Cartesian product-based tensor fusion layer to explicitly model intra-modality and inter-modality dynamics in an end-to-end manner.

\textbf{LMF.} Low-rank Multimodal Fusion (LMF) \cite{liu2018efficient} leverages modality-specific low-rank factors to compute tensor-based multimodal representations, which makes tensor fusion more efficient. 

\textbf{MulT.} Multimodal Transformer (MulT) \cite{tsai2019multimodal} utilizes directional pairwise cross-modal attention to capture inter-modal correlations in unaligned multimodal sequences. 

\textbf{Self-MM.} Self-Supervised Multi-task Multimodal (Self-MM) \cite{yu2021learning} sentiment analysis network [29] designs a unimodal label generation module based on self-supervised learning to explore unimodal supervision. 

\textbf{EMT.} Efficient Multimodal Transformer (EMT) \cite{sun2023efficient} with Dual-Level Feature Restoration for Robust Multimodal Sentiment Analysis.

\textbf{CMHFM.} A cross modal hierarchical fusion multimodal sentiment analysis method (CMHFM) \cite{wang2024cross} based on multi-task learning.

\subsection{Detection Performance}
To validate the performance of the models, we compared the state-of-the-art models as shown in Table \ref{table_dataset1}. The results of our methods are averaged over 5 runs. All other baseline performances are from published paper data. On the CH-SIMS dataset, our method outperforms all baselines on all metrics. The Acc-2 score improves by 0.6\%, the F1 score by 0.3\%, the Acc-3 score by 2.0\%, and the Acc-5 score by 3.4\% compared to the latest baseline model, CMHFM. The superior classification performance indicates that our MCIHN method is more effective than the baseline. Furthermore, the improvement in MAE and Corr indicates that the MCIHN method can better understand the CH-SIMS dataset. To further evaluate the effectiveness of our proposed model, we train the MCIHN model on the MOSI dataset. The results are shown in \ref{table_dataset1}, where our method improves the scores on Acc-2 and Acc-7 by 1.0\% and 0.8\% respectively. It also outperforms most of the baselines on the F1 and Corr metrics.

\subsection{Ablation Experiment}
To verify the validity of the MCIHN component and the validity of the three-modal fusion, we set up ablation experiments as follows: \textbf{-VT:} Model for visual and textual modal fusion. \textbf{-VA:} Model for visual and speech modal fusion. \textbf{-TA:} Model for textual and speech modal fusion. \textbf{MCIHN-1:} Remove the AAE module for multimodal fusion. \textbf{MCIHN-2:} Remove the Cross-modal Gate Mechanism Model (CGMM).

The experimental results in Table \ref{table_dataset2} show that the MCIHN model outperforms the MCIHN-1 and MCIHN-2 models. The MCIHN-2 model with the removal of the Cross-modal Gate Mechanism Model has the worst performance. The above results verify the effectiveness and feasibility of using the AAE module and the CGMM module for the emotion recognition task. The superiority of MCIHN mainly comes from the cross-modal gate mechanism. Compared with the model without this module, MCIHN will reduce the differences between different modalities and generate interactive modal features, which is conducive to the accuracy of classification after feature fusion. In addition, our model also satisfies the complex combination input of more modalities, which provides a valuable reference for future practical applications. Then, the experimental results from the bimodal input show that the text and visual features are better fused. However, the performance is still poor compared to the full MCIHN model, indicating that the strategy of three-modal fusion is optimal. The above experimental results on the missing modal model show that the text and visual information contain more emotional features that are easy to judge. At the same time, it also shows that the MCIHN model has a good balance effect on different modal inputs, and maximizes the extraction of useful features for classification.

\section{Conclusions}
This article introduces the MCIHN model, which is based on multi-path cross-modal interaction hybrid network. The MCIHN model has advantages over advanced models on the public datasets CH-SIMS and CMU-MOSI. In the future, we will continue to explore the validity of using video, voice, text and other features. Besides, we will further explore the correlated modal representations of sentiment and the interactions between core modalities and representation algorithms.


\bibliographystyle{ACM-Reference-Format}
\bibliography{refs}


\begin{thebibliography}{28}


\ifx \showCODEN    \undefined \def \showCODEN     #1{\unskip}     \fi
\ifx \showDOI      \undefined \def \showDOI       #1{#1}\fi
\ifx \showISBNx    \undefined \def \showISBNx     #1{\unskip}     \fi
\ifx \showISBNxiii \undefined \def \showISBNxiii  #1{\unskip}     \fi
\ifx \showISSN     \undefined \def \showISSN      #1{\unskip}     \fi
\ifx \showLCCN     \undefined \def \showLCCN      #1{\unskip}     \fi
\ifx \shownote     \undefined \def \shownote      #1{#1}          \fi
\ifx \showarticletitle \undefined \def \showarticletitle #1{#1}   \fi
\ifx \showURL      \undefined \def \showURL       {\relax}        \fi
\providecommand\bibfield[2]{#2}
\providecommand\bibinfo[2]{#2}
\providecommand\natexlab[1]{#1}
\providecommand\showeprint[2][]{arXiv:#2}

\bibitem[Dai et~al\mbox{.}(2021)]%
        {dai2021multimodal}
\bibfield{author}{\bibinfo{person}{Wenliang Dai}, \bibinfo{person}{Samuel
  Cahyawijaya}, \bibinfo{person}{Zihan Liu}, {and} \bibinfo{person}{Pascale
  Fung}.} \bibinfo{year}{2021}\natexlab{}.
\newblock \showarticletitle{Multimodal End-to-End Sparse Model for Emotion
  Recognition}. In \bibinfo{booktitle}{\emph{Proceedings of the 2021 Conference
  of the North American Chapter of the Association for Computational
  Linguistics: Human Language Technologies}}. \bibinfo{pages}{5305--5316}.
\newblock


\bibitem[Degottex et~al\mbox{.}(2014)]%
        {degottex2014covarep}
\bibfield{author}{\bibinfo{person}{Gilles Degottex}, \bibinfo{person}{John
  Kane}, \bibinfo{person}{Thomas Drugman}, \bibinfo{person}{Tuomo Raitio},
  {and} \bibinfo{person}{Stefan Scherer}.} \bibinfo{year}{2014}\natexlab{}.
\newblock \showarticletitle{COVAREP—A collaborative voice analysis repository
  for speech technologies}. In \bibinfo{booktitle}{\emph{2014 ieee
  international conference on acoustics, speech and signal processing
  (icassp)}}. IEEE, \bibinfo{pages}{960--964}.
\newblock


\bibitem[Fu et~al\mbox{.}(2024)]%
        {fu2024hybrid}
\bibfield{author}{\bibinfo{person}{Yanping Fu}, \bibinfo{person}{Zhiyuan
  Zhang}, \bibinfo{person}{Ruidi Yang}, {and} \bibinfo{person}{Cuiyou Yao}.}
  \bibinfo{year}{2024}\natexlab{}.
\newblock \showarticletitle{Hybrid cross-modal interaction learning for
  multimodal sentiment analysis}.
\newblock \bibinfo{journal}{\emph{Neurocomputing}}  \bibinfo{volume}{571}
  (\bibinfo{year}{2024}), \bibinfo{pages}{127201}.
\newblock


\bibitem[Hazarika et~al\mbox{.}(2020)]%
        {hazarika2020misa}
\bibfield{author}{\bibinfo{person}{Devamanyu Hazarika}, \bibinfo{person}{Roger
  Zimmermann}, {and} \bibinfo{person}{Soujanya Poria}.}
  \bibinfo{year}{2020}\natexlab{}.
\newblock \showarticletitle{Misa: Modality-invariant and-specific
  representations for multimodal sentiment analysis}. In
  \bibinfo{booktitle}{\emph{Proceedings of the 28th ACM international
  conference on multimedia}}. \bibinfo{pages}{1122--1131}.
\newblock


\bibitem[Huang et~al\mbox{.}(2017)]%
        {huang2017cross}
\bibfield{author}{\bibinfo{person}{Xin Huang}, \bibinfo{person}{Yuxin Peng},
  {and} \bibinfo{person}{Mingkuan Yuan}.} \bibinfo{year}{2017}\natexlab{}.
\newblock \showarticletitle{Cross-modal common representation learning by
  hybrid transfer network}. In \bibinfo{booktitle}{\emph{Proceedings of the
  26th International Joint Conference on Artificial Intelligence}}.
  \bibinfo{pages}{1893--1900}.
\newblock


\bibitem[Li et~al\mbox{.}(2019)]%
        {li2019towards}
\bibfield{author}{\bibinfo{person}{Runnan Li}, \bibinfo{person}{Zhiyong Wu},
  \bibinfo{person}{Jia Jia}, \bibinfo{person}{Yaohua Bu},
  \bibinfo{person}{Sheng Zhao}, {and} \bibinfo{person}{Helen Meng}.}
  \bibinfo{year}{2019}\natexlab{}.
\newblock \showarticletitle{Towards Discriminative Representation Learning for
  Speech Emotion Recognition.}. In \bibinfo{booktitle}{\emph{IJCAI}}.
  \bibinfo{pages}{5060--5066}.
\newblock


\bibitem[Liu and Shen(2018)]%
        {liu2018efficient}
\bibfield{author}{\bibinfo{person}{Zhun Liu} {and} \bibinfo{person}{Ying
  Shen}.} \bibinfo{year}{2018}\natexlab{}.
\newblock \showarticletitle{Efficient Low-rank Multimodal Fusion with
  Modality-Specific Factors}. In \bibinfo{booktitle}{\emph{Proceedings of the
  56th Annual Meeting of the Association for Computational Linguistics (Long
  Papers)}}.
\newblock


\bibitem[Mai et~al\mbox{.}(2020)]%
        {mai2020multi}
\bibfield{author}{\bibinfo{person}{Sijie Mai}, \bibinfo{person}{Haifeng Hu},
  \bibinfo{person}{Jia Xu}, {and} \bibinfo{person}{Songlong Xing}.}
  \bibinfo{year}{2020}\natexlab{}.
\newblock \showarticletitle{Multi-fusion residual memory network for multimodal
  human sentiment comprehension}.
\newblock \bibinfo{journal}{\emph{IEEE Transactions on Affective Computing}}
  \bibinfo{volume}{13}, \bibinfo{number}{1} (\bibinfo{year}{2020}),
  \bibinfo{pages}{320--334}.
\newblock


\bibitem[Makhzani(2018)]%
        {makhzani2018unsupervised}
\bibfield{author}{\bibinfo{person}{Alireza Makhzani}.}
  \bibinfo{year}{2018}\natexlab{}.
\newblock \bibinfo{booktitle}{\emph{Unsupervised representation learning with
  autoencoders}}.
\newblock \bibinfo{publisher}{University of Toronto (Canada)}.
\newblock


\bibitem[Mirsamadi et~al\mbox{.}(2017)]%
        {mirsamadi2017automatic}
\bibfield{author}{\bibinfo{person}{Seyedmahdad Mirsamadi},
  \bibinfo{person}{Emad Barsoum}, {and} \bibinfo{person}{Cha Zhang}.}
  \bibinfo{year}{2017}\natexlab{}.
\newblock \showarticletitle{Automatic speech emotion recognition using
  recurrent neural networks with local attention}. In
  \bibinfo{booktitle}{\emph{2017 IEEE International conference on acoustics,
  speech and signal processing (ICASSP)}}. IEEE, \bibinfo{pages}{2227--2231}.
\newblock


\bibitem[Pan et~al\mbox{.}(2020)]%
        {pan2020multi}
\bibfield{author}{\bibinfo{person}{Zexu Pan}, \bibinfo{person}{Zhaojie Luo},
  \bibinfo{person}{Jichen Yang}, {and} \bibinfo{person}{Haizhou Li}.}
  \bibinfo{year}{2020}\natexlab{}.
\newblock \showarticletitle{Multi-Modal Attention for Speech Emotion
  Recognition}. In \bibinfo{booktitle}{\emph{Proc. Interspeech 2020}}.
  \bibinfo{pages}{364--368}.
\newblock


\bibitem[Peng et~al\mbox{.}(2023)]%
        {peng2023fine}
\bibfield{author}{\bibinfo{person}{Junjie Peng}, \bibinfo{person}{Ting Wu},
  \bibinfo{person}{Wenqiang Zhang}, \bibinfo{person}{Feng Cheng},
  \bibinfo{person}{Shuhua Tan}, \bibinfo{person}{Fen Yi}, {and}
  \bibinfo{person}{Yansong Huang}.} \bibinfo{year}{2023}\natexlab{}.
\newblock \showarticletitle{A fine-grained modal label-based multi-stage
  network for multimodal sentiment analysis}.
\newblock \bibinfo{journal}{\emph{Expert Systems with Applications}}
  \bibinfo{volume}{221} (\bibinfo{year}{2023}), \bibinfo{pages}{119721}.
\newblock


\bibitem[Rahman et~al\mbox{.}(2020)]%
        {rahman2020integrating}
\bibfield{author}{\bibinfo{person}{Wasifur Rahman}, \bibinfo{person}{Md~Kamrul
  Hasan}, \bibinfo{person}{Sangwu Lee}, \bibinfo{person}{Amir Zadeh},
  \bibinfo{person}{Chengfeng Mao}, \bibinfo{person}{Louis-Philippe Morency},
  {and} \bibinfo{person}{Ehsan Hoque}.} \bibinfo{year}{2020}\natexlab{}.
\newblock \showarticletitle{Integrating multimodal information in large
  pretrained transformers}. In \bibinfo{booktitle}{\emph{Proceedings of the
  conference. Association for Computational Linguistics. Meeting}},
  Vol.~\bibinfo{volume}{2020}. NIH Public Access, \bibinfo{pages}{2359}.
\newblock


\bibitem[Sun et~al\mbox{.}(2023a)]%
        {sun2023using}
\bibfield{author}{\bibinfo{person}{Dekai Sun}, \bibinfo{person}{Yancheng He},
  {and} \bibinfo{person}{Jiqing Han}.} \bibinfo{year}{2023}\natexlab{a}.
\newblock \showarticletitle{Using auxiliary tasks in multimodal fusion of
  wav2vec 2.0 and bert for multimodal emotion recognition}. In
  \bibinfo{booktitle}{\emph{ICASSP 2023-2023 IEEE International Conference on
  Acoustics, Speech and Signal Processing (ICASSP)}}. IEEE,
  \bibinfo{pages}{1--5}.
\newblock


\bibitem[Sun et~al\mbox{.}(2024)]%
        {sun2024fine}
\bibfield{author}{\bibinfo{person}{Haoqin Sun}, \bibinfo{person}{Shiwan Zhao},
  \bibinfo{person}{Xuechen Wang}, \bibinfo{person}{Wenjia Zeng},
  \bibinfo{person}{Yong Chen}, {and} \bibinfo{person}{Yong Qin}.}
  \bibinfo{year}{2024}\natexlab{}.
\newblock \showarticletitle{Fine-Grained Disentangled Representation Learning
  For Multimodal Emotion Recognition}. In \bibinfo{booktitle}{\emph{ICASSP
  2024-2024 IEEE International Conference on Acoustics, Speech and Signal
  Processing (ICASSP)}}. IEEE, \bibinfo{pages}{11051--11055}.
\newblock


\bibitem[Sun et~al\mbox{.}(2023b)]%
        {sun2023efficient}
\bibfield{author}{\bibinfo{person}{Licai Sun}, \bibinfo{person}{Zheng Lian},
  \bibinfo{person}{Bin Liu}, {and} \bibinfo{person}{Jianhua Tao}.}
  \bibinfo{year}{2023}\natexlab{b}.
\newblock \showarticletitle{Efficient multimodal transformer with dual-level
  feature restoration for robust multimodal sentiment analysis}.
\newblock \bibinfo{journal}{\emph{IEEE Transactions on Affective Computing}}
  \bibinfo{volume}{15}, \bibinfo{number}{1} (\bibinfo{year}{2023}),
  \bibinfo{pages}{309--325}.
\newblock


\bibitem[Tao and Tan(2009)]%
        {tao2009affective}
\bibfield{author}{\bibinfo{person}{Jianhua Tao} {and} \bibinfo{person}{Tieniu
  Tan}.} \bibinfo{year}{2009}\natexlab{}.
\newblock \bibinfo{booktitle}{\emph{Affective information processing}}.
\newblock \bibinfo{publisher}{Springer}.
\newblock


\bibitem[Tsai et~al\mbox{.}(2019)]%
        {tsai2019multimodal}
\bibfield{author}{\bibinfo{person}{Yao-Hung~Hubert Tsai},
  \bibinfo{person}{Shaojie Bai}, \bibinfo{person}{Paul~Pu Liang},
  \bibinfo{person}{J~Zico Kolter}, \bibinfo{person}{Louis-Philippe Morency},
  {and} \bibinfo{person}{Ruslan Salakhutdinov}.}
  \bibinfo{year}{2019}\natexlab{}.
\newblock \showarticletitle{Multimodal transformer for unaligned multimodal
  language sequences}. In \bibinfo{booktitle}{\emph{Proceedings of the
  conference. Association for computational linguistics. Meeting}},
  Vol.~\bibinfo{volume}{2019}. NIH Public Access, \bibinfo{pages}{6558}.
\newblock


\bibitem[Wang et~al\mbox{.}(2023b)]%
        {wang2023cross}
\bibfield{author}{\bibinfo{person}{Di Wang}, \bibinfo{person}{Shuai Liu},
  \bibinfo{person}{Quan Wang}, \bibinfo{person}{Yumin Tian},
  \bibinfo{person}{Lihuo He}, {and} \bibinfo{person}{Xinbo Gao}.}
  \bibinfo{year}{2023}\natexlab{b}.
\newblock \showarticletitle{Cross-Modal Enhancement Network for Multimodal
  Sentiment Analysis}.
\newblock \bibinfo{journal}{\emph{IEEE Transactions on Multimedia}}
  \bibinfo{volume}{25} (\bibinfo{year}{2023}), \bibinfo{pages}{4909--4921}.
\newblock


\bibitem[Wang et~al\mbox{.}(2023a)]%
        {wang2023multimodal}
\bibfield{author}{\bibinfo{person}{Fanfan Wang}, \bibinfo{person}{Zixiang
  Ding}, \bibinfo{person}{Rui Xia}, \bibinfo{person}{Zhaoyu Li}, {and}
  \bibinfo{person}{Jianfei Yu}.} \bibinfo{year}{2023}\natexlab{a}.
\newblock \showarticletitle{Multimodal Emotion-Cause Pair Extraction in
  Conversations}.
\newblock \bibinfo{journal}{\emph{IEEE Transactions on Affective Computing}}
  \bibinfo{volume}{14}, \bibinfo{number}{3} (\bibinfo{year}{2023}),
  \bibinfo{pages}{1832--1844}.
\newblock


\bibitem[Wang et~al\mbox{.}(2024)]%
        {wang2024cross}
\bibfield{author}{\bibinfo{person}{Lan Wang}, \bibinfo{person}{Junjie Peng},
  \bibinfo{person}{Cangzhi Zheng}, \bibinfo{person}{Tong Zhao},
  {et~al\mbox{.}}} \bibinfo{year}{2024}\natexlab{}.
\newblock \showarticletitle{A cross modal hierarchical fusion multimodal
  sentiment analysis method based on multi-task learning}.
\newblock \bibinfo{journal}{\emph{Information Processing \& Management}}
  \bibinfo{volume}{61}, \bibinfo{number}{3} (\bibinfo{year}{2024}),
  \bibinfo{pages}{103675}.
\newblock


\bibitem[Yang et~al\mbox{.}(2023)]%
        {yang2023behavioral}
\bibfield{author}{\bibinfo{person}{Kangning Yang}, \bibinfo{person}{Chaofan
  Wang}, \bibinfo{person}{Yue Gu}, \bibinfo{person}{Zhanna Sarsenbayeva},
  \bibinfo{person}{Benjamin Tag}, \bibinfo{person}{Tilman Dingler},
  \bibinfo{person}{Greg Wadley}, {and} \bibinfo{person}{Jorge Goncalves}.}
  \bibinfo{year}{2023}\natexlab{}.
\newblock \showarticletitle{Behavioral and Physiological Signals-Based Deep
  Multimodal Approach for Mobile Emotion Recognition}.
\newblock \bibinfo{journal}{\emph{IEEE Transactions on Affective Computing}}
  \bibinfo{volume}{14}, \bibinfo{number}{02} (\bibinfo{year}{2023}),
  \bibinfo{pages}{1082--1097}.
\newblock


\bibitem[Yu et~al\mbox{.}(2020)]%
        {yu2020ch}
\bibfield{author}{\bibinfo{person}{Wenmeng Yu}, \bibinfo{person}{Hua Xu},
  \bibinfo{person}{Fanyang Meng}, \bibinfo{person}{Yilin Zhu},
  \bibinfo{person}{Yixiao Ma}, \bibinfo{person}{Jiele Wu},
  \bibinfo{person}{Jiyun Zou}, {and} \bibinfo{person}{Kaicheng Yang}.}
  \bibinfo{year}{2020}\natexlab{}.
\newblock \showarticletitle{Ch-sims: A chinese multimodal sentiment analysis
  dataset with fine-grained annotation of modality}. In
  \bibinfo{booktitle}{\emph{Proceedings of the 58th annual meeting of the
  association for computational linguistics}}. \bibinfo{pages}{3718--3727}.
\newblock


\bibitem[Yu et~al\mbox{.}(2021)]%
        {yu2021learning}
\bibfield{author}{\bibinfo{person}{Wenmeng Yu}, \bibinfo{person}{Hua Xu},
  \bibinfo{person}{Ziqi Yuan}, {and} \bibinfo{person}{Jiele Wu}.}
  \bibinfo{year}{2021}\natexlab{}.
\newblock \showarticletitle{Learning modality-specific representations with
  self-supervised multi-task learning for multimodal sentiment analysis}. In
  \bibinfo{booktitle}{\emph{Proceedings of the AAAI conference on artificial
  intelligence}}, Vol.~\bibinfo{volume}{35}. \bibinfo{pages}{10790--10797}.
\newblock


\bibitem[Zadeh et~al\mbox{.}(2017)]%
        {zadeh2017tensor}
\bibfield{author}{\bibinfo{person}{Amir Zadeh}, \bibinfo{person}{Minghai Chen},
  \bibinfo{person}{Soujanya Poria}, \bibinfo{person}{Erik Cambria}, {and}
  \bibinfo{person}{Louis-Philippe Morency}.} \bibinfo{year}{2017}\natexlab{}.
\newblock \showarticletitle{Tensor Fusion Network for Multimodal Sentiment
  Analysis}. In \bibinfo{booktitle}{\emph{Proceedings of the 2017 Conference on
  Empirical Methods in Natural Language Processing}}.
  \bibinfo{pages}{1103--1114}.
\newblock


\bibitem[Zadeh et~al\mbox{.}(2016)]%
        {zadeh2016multimodal}
\bibfield{author}{\bibinfo{person}{Amir Zadeh}, \bibinfo{person}{Rowan
  Zellers}, \bibinfo{person}{Eli Pincus}, {and} \bibinfo{person}{Louis-Philippe
  Morency}.} \bibinfo{year}{2016}\natexlab{}.
\newblock \showarticletitle{Multimodal sentiment intensity analysis in videos:
  Facial gestures and verbal messages}.
\newblock \bibinfo{journal}{\emph{IEEE Intelligent Systems}}
  \bibinfo{volume}{31}, \bibinfo{number}{6} (\bibinfo{year}{2016}),
  \bibinfo{pages}{82--88}.
\newblock


\bibitem[Zhang et~al\mbox{.}(2023)]%
        {zhang2023deep}
\bibfield{author}{\bibinfo{person}{Shiqing Zhang}, \bibinfo{person}{Yijiao
  Yang}, \bibinfo{person}{Chen Chen}, \bibinfo{person}{Xingnan Zhang},
  \bibinfo{person}{Qingming Leng}, {and} \bibinfo{person}{Xiaoming Zhao}.}
  \bibinfo{year}{2023}\natexlab{}.
\newblock \showarticletitle{Deep learning-based multimodal emotion recognition
  from audio, visual, and text modalities: A systematic review of recent
  advancements and future prospects}.
\newblock \bibinfo{journal}{\emph{Expert Systems with Applications}}
  (\bibinfo{year}{2023}), \bibinfo{pages}{121692}.
\newblock


\bibitem[Zheng et~al\mbox{.}(2023)]%
        {zheng2023multi}
\bibfield{author}{\bibinfo{person}{Jiahao Zheng}, \bibinfo{person}{Sen Zhang},
  \bibinfo{person}{Zilu Wang}, \bibinfo{person}{Xiaoping Wang}, {and}
  \bibinfo{person}{Zhigang Zeng}.} \bibinfo{year}{2023}\natexlab{}.
\newblock \showarticletitle{Multi-Channel Weight-Sharing Autoencoder Based on
  Cascade Multi-Head Attention for Multimodal Emotion Recognition}.
\newblock \bibinfo{journal}{\emph{IEEE Transactions on Multimedia}}
  \bibinfo{volume}{25} (\bibinfo{year}{2023}), \bibinfo{pages}{2213--2225}.
\newblock


\end{thebibliography}


\end{document}